\documentclass[11pt,a4paper]{article}
\usepackage[hyperref]{naaclhlt2018}
\usepackage{times}
\usepackage{CJK}

\usepackage{helvet}
\usepackage{courier}
\usepackage{graphicx}
\usepackage{multirow}
\usepackage{amsmath}

\newcommand*{\affaddr}[1]{#1} 
\newcommand*{\affmark}[1][*]{\textsuperscript{#1}}
\newcommand*{\email}[1]{\texttt{#1}}

\usepackage{subcaption}
\usepackage{tikz-dependency}
\usepackage{tikz-qtree}

\usepackage{latexsym}

\usepackage{url}

\aclfinalcopy 

\title{Structure Regularized Neural Network for Entity Relation Classification for Chinese Literature Text}

\author{Ji Wen\affmark[1], Xu Sun\affmark[1,2], Xuancheng Ren\affmark[1], Qi Su\affmark[3]\\
\affaddr{\affmark[1]MOE Key Lab of Computational Linguistics, School of EECS, Peking University}\\
\affaddr{\affmark[2]Deep Learning Lab, Beijing Institute of Big Data Research, Peking University}\\
\affaddr{\affmark[3]School of Foreign Languages, Peking University}\\
\email{\{wenjics, xusun, renxc, sukia\}@pku.edu.cn}\\
}

\date{}

\begin{document}
\begin{CJK*}{UTF8}{gbsn}

\maketitle
\begin{abstract}
Relation classification is an important semantic processing task in the field of natural language processing. In this paper, we propose the task of relation classification for Chinese literature text. A new dataset of Chinese literature text is constructed to facilitate the study in this task. We present a novel model, named Structure Regularized Bidirectional Recurrent Convolutional Neural Network (SR-BRCNN), to identify the relation between entities. 
The proposed model learns relation representations along the shortest dependency path (SDP) extracted from the structure regularized dependency tree, which has the benefits of reducing the complexity of the whole model. Experimental results show that the proposed method significantly improves the $F_{1}$ score by 10.3, and outperforms the state-of-the-art approaches on Chinese literature text\footnote{The Chinese literature text corpus, which this paper developed and used, is available at \url{https://github.com/lancopku/Chinese-Literature-NER-RE-Dataset}.}.

\end{abstract}

\section{Introduction}

Relation classification is the task of identifying the semantic relation holding between two nominal entities in text. 
Recently, neural networks are widely used in relation classification. \citet{wang2016relation} proposes a convolutional neural network with two levels of attention. \citet{zhang2015bidirectional} uses bidirectional long short-term memory networks to model the sentence with sequential information.
\citet{bunescu-mooney2005} first uses SDP between two entities to capture the predicate-argument sequences. 
\citet{DBLP:conf/ijcnlp/WangLYSW17} explores the idea of incorporating syntactic
parse tree into neural networks.
\citet{liu2017soft} proposes a noise-tolerant method to deal with wrong labels in distant-supervised relation extraction with soft labels.
In recent years, we have seen a move towards deep learning architectures. \citet{liu2015dependency} develops dependency-based neural networks. \citet{xu2015classifying} applies long short term memory (LSTM) \cite{hochreiter1997long} based recurrent neural networks (RNNs) along with the SDP. 
 
\begin{figure*} 
	\centering
	\centerline{\includegraphics[width= 0.7\linewidth]{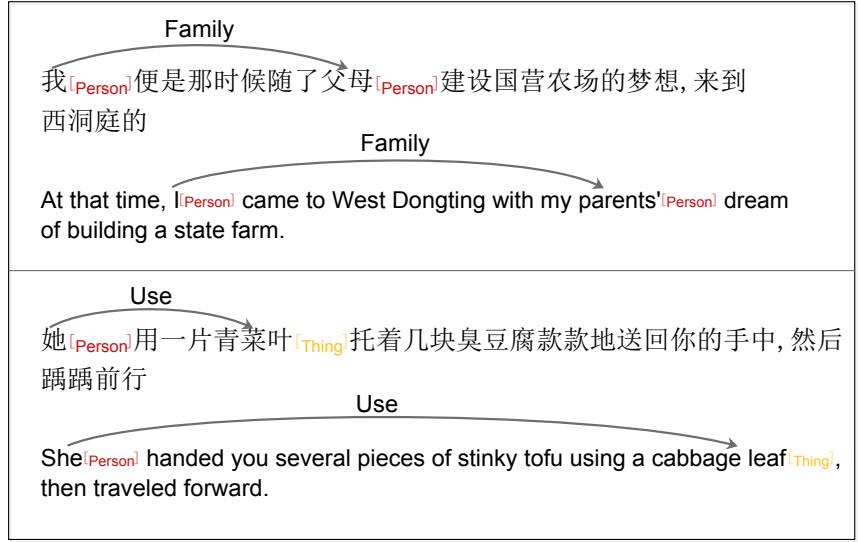}}
	\caption{Examples from the Chinese literature text corpus.}\label{fig1}
\end{figure*}

In this paper, we focus on relation classification of Chinese literature texts, which to our knowledge has not been studied before, due to the challenge that it brings. 
Chinese literature text tends to express intuitions and feelings rather than presenting scientific analysis. 
It has a wide range of topics. Many literature articles express feelings in a subtle and special way, making it more difficult to recognize entities. 
Chinese literature text is not organized very logically, whether among paragraphs or sentences. They tend to use various forms of sentences to create free feelings. 
Sentence structure of them is very flexible. The sentences are not associated with each other by evident conjunctions. Besides, Chinese is a topic-prominent language, the subject is usually covert and the usage of words is relatively flexible.

In short, sentences of Chinese literature text contain many non-essential words, and embody very complex and flexible structures. Existing methods make intensive use of the syntactical information, such as part-of-speech tags, and dependency relations. However, the automatically generated information is not reliable and of poor quality for Chinese literature text. It is of great challenge for the existing methods to achieve satisfying performance.

To mitigate the noisy syntactical information, we propose to apply structure regularization to the structures used in relation classification.
Structured prediction models are commonly used to solve structure dependent problems in a wide variety of application domains. Recently, many existing systems on structured prediction focus on increasing the level of structural dependencies within the model. However, the theoretical and experimental study of \citet{sun2014structure} suggests that complex structures are tending to increase the overfitting risk, and can potentially be harmful to the model accuracy. As pointed out by \citet{sun2014structure}, complex structural dependencies have a drawback of increasing the generalization risk, because more complex structures are easier to suffer from overfitting.

In this paper, we focus on the study of applying structure regularization to the relation classification task of Chinese literature text.  To summarize, the contributions of this paper are as follows:

\begin{itemize}
	\item To our knowledge, we are the first to develop a corpus of Chinese literature text for relation classification. The corpus contains 837 articles. It helps alleviate the dilemma of the lack of corpus in Chinese Relation Classification.
    
	\item We develop the tree-based structure regularization methods and make progress on the task of relation classification.  The method of structure regularization is normally used on the structure of sequences, while we find a way to realize it on the structure of trees. Comparing to the origin model, applying structure regularization substantially improves the $F_{1}$ score by 10.3.
	
\end{itemize}

\section{Chinese Literature Text Corpus}

\begin{CJK*}{UTF8}{gbsn}
\begin{table*}[!hbt]
\centering

\begin{tabular}{cccc}
\hline
Tag&Description&Example&\%\\
\hline
Located&locate in&  幽兰(orchid)-山谷(valley)&37.43\\

\hline
Part-Whole&be a part of&  花(flower)-仙人掌(cactus)&23.76\\
\hline
Family&be family members&  母亲(mother)-奶奶(grandmother)&10.25\\

\hline
General-Special&be a general range and a special kind of it&  鱼(fish)-鲫鱼(carp)&6.99\\

\hline
Social&be socially related&  母亲(mother)-邻里(neighbour)&6.02\\

\hline
Ownership&be in possession of&  村民(villager)-旧屋(house)&5.10\\

\hline
Use&do something with&  爷爷(grandfather)-毛笔(brush)&4.76\\

\hline
Create&make happen or exist&  男人(man)-陶器(pottery)&2.93\\

\hline
Near&a short distance away&  山(hill)-县城(town)&2.76\\

\hline
\end{tabular}
\caption{The set of relation tags. The
last column indicates each tag's relative frequency in the
full annotated data. }
\label{relationTypes}
\end{table*}
\end{CJK*}

In Figure~\ref{fig1}, we show two examples from the annotated corpus. We label the entities and relations of the text on a sentence level. There are 6 kinds of entities and 9 kinds of relations. The task aims at predicting the labels of these relations, given the sentences as well as the entities and their types. The corpus is part of the work of \citet{DBLP:journals/corr/abs-1711-07010}.

We obtain over 1,000 Chinese prose articles from the website and then filter and extract 837 articles. Too short or too noise articles are not included. Due to the difficulty of tagging Chinese prose texts, We divide the annotation process into three steps. 

First, we attempt to annotate the raw articles based on defined entity and relation tags. In the tagging process, we find a problem of data inconsistency. To solve this problem, we design the next two steps. 
Second, we design several generic disambiguation rules to ensure the consistency of annotation guidelines. For example, remove all adjective words and only tag ``entity header'' when tagging entities (e.g., change ``a girl in red cloth'' to ``girl''). In this stage, we re-annotate all articles and correct all inconsistency entities based on the heuristic rules.
Even though the heuristic tagging process significantly improves dataset quality, it is too hard to handle all inconsistency cases based on limited heuristic rules. Finally, we introduce a machine auxiliary tagging method. The core idea is to train a model to learn annotation guidelines on the subset of the corpus and produce predicted tags on the rest data. The predicted tags are used to be compared with the gold tags to discovery inconsistent entities, which largely reduce annotators' efforts.  
After all annotation steps, we also manually check all entities and relations to ensure the correctness of corpus.

We set 9 different classes for better understanding the connection between pairs of entities of proses, including ``Located'', ``Near'', ``Part-Whole'', ``Family'', ``Social'', ``Create'', ``Use'', ``Ownership'', and ``General-Special''. 
Details of the tags are shown in \ref{relationTypes}.
For building the relations between people in proses, we use the ``Social'' tag, which is not quite common in other corpora. At the same time, the ``Family'' tag represents the relation between family members, which is a special kind of relation of the ``Social'' tag. It is because the ``Family'' relation is a much stronger bond than the ``Social'' relation and the number of the entity pairs tagged as ``Family'' is almost two time of that tagged as ``Social''.

In prior work, Chinese literature text corpus is very rare. Many tasks cannot achieve a satisfying result on Chinese literature text compared to other corpus. However, understanding Chinese literature text is of great importance to Chinese literature research.

\begin{table*}[t]
\centering
\begin{tabular}{c|c|c|c}
\hline
\multicolumn{1}{c|}{\multirow{1}{*}{}}&\multicolumn{1}{c|}{\multirow{1}{*}{Models}}&Information&\multicolumn{1}{c}{\multirow{1}{*}{ $F_{1}$ }}\\
\hline
\multicolumn{1}{c|}{\multirow{4}{*}{\textbf{Baselines}}}
&SVM&Word embeddings, NER, WordNet, HowNet,  &48.9\\
&\cite{Hendrickx2010}&POS, dependency parse, Google n-gram&\\
\cline{2-4}
&RNN&Word embeddings&48.3\\
&\cite{socher2011semi}&+ POS, NER, WordNet&49.1\\
\cline{2-4}
&CNN&Word embeddings&47.6\\
&\cite{zeng2014relation}&+ word position embeddings, NER, WordNet&52.4\\
\cline{2-4}
&CR-CNN&Word embeddings&52.7\\
&\cite{santos2015classifying}&+ word position embeddings&54.1\\
\cline{2-4}
&SDP-LSTM&Word embeddings&54.9\\
&\cite{xu2015classifying}&+ POS + NER + WordNet&55.3\\
\cline{2-4}
&DepNN&Word embeddings, WordNet&55.2\\
&\cite{liu2015dependency}&&\\
\cline{2-4}
&BRCNN&Word embeddings&55.0\\
&\cite{cai2016bidirectional}&+ POS, NER, WordNet&55.6\\
\hline

\hline
\multicolumn{1}{c|}{\multirow{2}{*}{\textbf{Our Model}}}

& \textbf{SR-BRCNN} &Word embeddings& \textbf{65.2 (+9.6)} \\
&&+ POS, NER, WordNet& \textbf{65.9 (+10.3)} \\

\cline{2-4}
\hline
\end{tabular}
\caption{Comparison of relation classification systems on Chinese literature text.}\label{daimprovements}
\end{table*}

\section{Structure Regularized BRCNN}

\subsection{Basic BRCNN}


The Bidirectional Recurrent Convolutional Neural Network (BRCNN) model is used to learn representations with bidirectional information along the shortest dependency path (SDP). 

Given a sentence and its dependency tree, we build our neural network on its SDP extracted from tree. Along the SDP, recurrent neural networks with long short term memory units are applied to learn hidden representations of words and dependency relations, respectively. A convolution layer is applied to capture local features from hidden representations of every two neighbor words and the dependency relations between them. A max pooling layer thereafter gathers information from local features of the SDP and the inverse SDP. We have a softmax output layer after pooling layer for classification in the unidirectional model RCNN.

On the basis of RCNN model, we build a bidirectional architecture BRCNN taking the SDP and the inverse SDP of a sentence as input. During the training stage of a (K+1)-relation task, two fine-grained softmax classifiers of RCNNs do a (2K + 1)-class classification respectively. The pooling layers of two RCNNs are concatenated and a coarse-grained softmax output layer is followed to do a (K + 1)-class classification. The final (2K+1)-class distribution is the combination of two (2K+1)-class distributions provided by fine grained classifiers during the testing stage.

\citet{bunescu-mooney2005} first uses shortest dependency paths between two entities to capture the predicate-argument sequences, which provided strong evidence for relation classification. Each two neighbor words are linked by a dependency relation in shortest dependency path. The order of the words will affect the meaning of relations. Single direction of relation may not reflect all information in context. Thus, we employ a bidirectional recurrent convolutional neural network to capture more information from the sentence. The corresponding relation keeps the same when we inverse the shortest dependency path.

Due to the limitations of recurrent neural networks to capture longterm dependencies, we employ LSTM in our work. LSTM performs better in tasks where long dependencies are needed. Some gating units are designed in a LSTM cell. Each of them are in charge of specific functions. We use two bidirectional LSTMs to capture the features of words and relations separately. Word embedding and relation embedding are initialized with two look up tables. Then we can get a real-valued vector of every word and relation according to their index. In recurrent neural networks, the input is the current embedding $x_t$ and it previous state $h_{t-1}$. For the LSTM that captures word information, $x_t$ is the word embedding and for the LSTM that captures relation information, $x_t$ is the relation embedding. The current step output is denoted as $h_t$. We consider it a representation of all information until this time step. Moreover, a bidirectional LSTM is used to capture the previous and later information, as we did in this work.

Convolutional neural network performs well in capturing local features. After we obtain representations of words and relations, we concatenate them to get a representation of a complete dependency unit. The hidden state of a relation is denoted as $r_{ab}$. Words on its sides have the hidden states denoted as $h_a$ and $h_b$.  [$h_a$ $h_{ab}$ $h_b$] denotes the representation of a dependency unit $L_{ab}$. Then we utilize a convolution layer upon the concatenation. We have
\begin{equation}
L_{ab} = f(W_{con} \cdot [h_a \oplus h'_{ab} \oplus h_b] + b_{con})
\end{equation}
where $W_{con}$ is the weight matrix and $b_{con}$ is a bias term. We choose $tanh$ as our activation function and a max pooling followed.

Recurrent neural networks have a long memory, while it causes a distance bias problem. Where inputs are exactly the same may have different representations due to the position in a sentence. However, entities and key components could appear anywhere in a SDP. Thus, two RCNNs pick up information along the SDP and it reverse. A coarse-grained softmax classifier is applied on the global representations $\overrightarrow{G}$ and $ \overleftarrow{G}$. Two fine-grained softmax classifier are applied to to give a more detailed prediction of (2K+1) class.
\begin{equation}
\overrightarrow{y} = softmax(W_{f} \cdot \overrightarrow{G} + b_{f}) 
\end{equation}
\begin{equation}
\overleftarrow{y} = softmax(W_{f} \cdot \overleftarrow{G} + b_{f})
\end{equation}

During training, our objective is the penalized cross-entropy of three classifiers. Formally,
\begin{equation}
\begin{split}
J = &\sum_{i=1}^{2K+1}\overrightarrow{t_{i}}log\overrightarrow{y_{i}}+\sum_{i=1}^{2K+1}\overleftarrow{t_{i}}log\overleftarrow{y_{i}} \\
&+\sum_{i=1}^{K}{t_{i}}logy_{i}+\lambda \cdot \Arrowvert \theta \Arrowvert^{2}
\end{split}
\end{equation}
When decoding, the final prediction is a combination of $\overrightarrow{y}$ and  $\overleftarrow{y}$
\begin{equation}
y_{test} = \alpha \cdot\overrightarrow{y} + (1 - \alpha) \cdot z(\overleftarrow{y})
\end{equation}

\subsection{Structure Regularized BRCNN}
The basic BRCNN model can handle the task well, but there still remains some weakness, especially dealing with long sentences with complicated structures. The SDP generated from a more complicated dependency tree consists more irrelevant words. \citet{DBLP:journals/corr/Sun14c} shows both theoretically and empirically that structure regularization can effectively control overfitting risk and lead to better performance. \citet{DBLP:journals/corr/abs-1711-06528} and \citet{DBLP:journals/corr/abs-1711-10331} also show that complex structure models are prone to the structure-based overfitting. 
Therefore, we propose the structure regularized BRCNN.
  
We conduct structure regularization on the dependency tree of the sentences. Based on the heuristic rules, several nodes in the dependency tree are selected. The subtrees of these selected nodes are cut from the whole dependency tree. With these selected nodes as the roots, these subtrees form a forest. The forest will be connected by lining the roots of the trees of the forest. Traditional SDP is extracted directly from the dependency tree, while in our model, the SDP is extracted from the final forest. We call these kinds of SDPs as SR-SDPs. We build our BRCNN model on the SR-SDP. 

\begin{figure}[t]
    \centering
    \subcaptionbox{The dependency tree and the SDP before flattening.\label{fig:ex-1}}{
\begin{tikzpicture}[
	level 1/.style={sibling distance = 1cm, level distance = 1cm},
    level 2/.style={sibling distance = 1cm},
    level 3/.style={sibling distance = 1cm},
    edge from parent/.style={->,draw}
    ]
	\node (is-root) {a}
		child { node [circle,draw] {\textbf{b}} edge from parent[thick,draw] }
		child {
			node {c} edge from parent[thick,draw]
				child { node {d} edge from parent[thin,draw] }
				child { 
                	node {e} edge from parent[thick,draw]
                		child {node {f} edge from parent[thin,draw]}
                     	child {node [circle,draw,thin] {\textbf{g}} edge from parent[thick,draw]}
                }
		};
\end{tikzpicture}
        }
    \quad
    \subcaptionbox{The dependency tree and the SDP after flattening.\label{fig:ex-2}}{
        \begin{tikzpicture}[
	level 1/.style={sibling distance = 1cm, level distance = 1cm},
    level 2/.style={sibling distance = 1cm},
    level 3/.style={sibling distance = 1cm},
    edge from parent/.style={->,draw}
    ]
	\node (is-root) {a}
		child { node [circle,draw] {\textbf{b}} edge from parent[thick,draw]}
		child {
			node {c} edge from parent[thin,draw]
				child { node {d} edge from parent[thin,draw] }
        }
        child{
               node {e} edge from parent[thick,draw]
                		child {node {f} edge from parent[thin,draw]}
                     	child {node [circle,draw,thin] {\textbf{g}} edge from parent[thick,draw] }
		};
\end{tikzpicture}
        }
    \caption{An example of the proposed method. The two words in circles are the entities, and the thick edges form the SDP. By flattening the dependency tree, the path becomes shorter.}
    \label{fig:ex}
\end{figure}
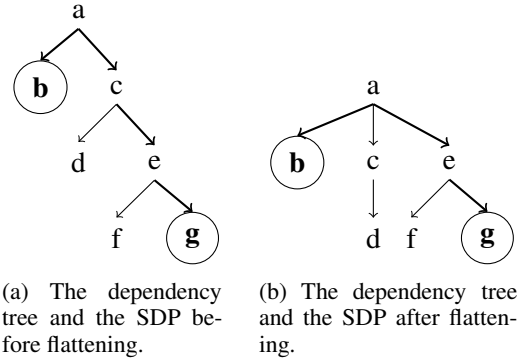

\subsection{Various Structure Regularization Methods}

We experiment with three kinds of regularization rules. First, The punctuation is a natural break point of the sentence. The resulting subtrees usually keep similar syntax to traditional dependency trees. Another popular method to regularize the structure is to decompose the structure randomly. In our model, we randomly select several nodes in the dependency tree and then cut the subtrees under these nodes. Finally we decide to cut the dependency tree by prepositions, especially in Chinese literature text. There usually are many decorations to describe the entities, and the using of prepositional phrases is very common for that purpose. So we also try to decompose the dependency trees using prepositions.

\section{Experiments}
We evaluated our model on the Chinese literature text corpus. It contains 9 distinguished types of relations among 837 articles. The dataset contains 695 articles for training,  58 for validation, and 84 for testing.

\subsection{Experiment settings}
We use pre-trained word embeddings, which are trained on Gigaword with word2vec \cite{mikolov2013distributed}. Word embeddings are 200-dimensional. The embeddings of relation are initialized randomly and are 50-dimensional. The hidden layers of LSTMs to extract information from entities and relations are the same as the embedding dimension of entities and relations. 
We applied L2 regularization to weights in neural networks and dropout to embeddings with a keep probability 0.5. AdaDelta \cite{zeiler2012adadelta} is used for optimization.

\subsection{Experimental Results}
Table~\ref{daimprovements} compares our SR-BRCNN model with other state-of-the-art methods on the corpus of Chinese literature text, including the basic BRCNN, \citet{cai2016bidirectional} method. Structure regularization helps improve the result a lot. The method of structure regularization could prevent the overfitting of poor quality SDPs.

\subsection{Analysis: Effect of SR}

Figure~\ref{fig:ex-1} and Figure~\ref{fig:ex-2} show an example of structure regularized SDP. The relation is between the two circled elements.   
The main idea of the method is to avoid the incorrect structure from the dependency trees generated by the parser. The SDP in Figure~\ref{fig:ex-1} is longer than the SR-SDP in Figure~\ref{fig:ex-2}. However, the dependency tree of the example is not completely correct. The longer the SDP is, the more incorrect information the model learns.

The structure regularized BRCNN has shown obvious improvements on both English and Chinese datasets. We attribute the improvements to the simplified structures that generated by structure regularization. The internal relations of components of a sentence are more obscure due to the feature of Chinese literature text. By conducting structure regularization on the dependency tree, we get several subtrees with simpler structure, and then we extract SDP from the lined forests. In most cases, the distance between two entities will be shortened along the new SR-SDP. Without the redundant information along the original SDP. The model that benefits from the intensive dependencies will capture more effective information for classification.

\subsection{Analysis: Effect of Different Regularization Methods}

\begin{table}[h]
		\centering
		\begin{tabular}{|l|c|}\hline
			Classifier&$F_{1}$ score\\\hline
			BRCNN&55.6\\
			SR by punctuation &59.7\\
			SR by random &62.4\\
			SR by preposition &65.9\\
			\hline
		\end{tabular}
		\caption{Different structure regularization results on Chinese literature texts.}
		\label{tab:Margin_settings}
	\end{table} 

The punctuation is a natural break point of the sentence, which makes subtrees more like the traditional dependency trees in the aspect of integrity. However, the original dependency trees cannot be sufficiently regularized. Despite its drawbacks, this method still shows obvious improvements on the model and leads to further experiments.

Regularizing the structure by decomposing the structure randomly will solve the insufficient decomposition problems. The method of structure regularization has shown that the degree of loss of information is not a serious problem. It gives a slightly better result compared to cutting dependency trees by punctuation.

A more elaborate method is to cut the dependency tree by prepositions. In Chinese literature text, prepositional phrases are used frequently. Cutting by prepositions will regularize the tree more sufficiently. Meanwhile, the subtrees under the prepositional nodes are usually internally linked.

\section{Conclusions}

In this paper, we present a novel model, Structure Regularized BRCNN, to classify the relation of two entities in a sentence. We demonstrate that tree-based structure regularization can help improve the results, while the method is normally used in sequence-based models before. The proposed structure regularization method  makes the SDP shorter and contains less noise from the unreliable parse trees. This leads to substantial improvement on the relation classification results. The results also show how different ways of regularization act in the model of BRCNN. The best way of them helps improve the $F_{1}$ score by 10.3.

We also develop a corpus on Chinese literature text focusing on the task of Relation Classification. The new corpus is large enough for us to train models and verify the models. 

\section*{Acknowledgements}

This work was supported in part by National Natural Science Foundation of China (No. 61673028), National High Technology Research and Development Program of China (863 Program, No. 2015AA015404), and the National Thousand Young Talents Program. Xu Sun is the corresponding author of this paper.

\nocite{DBLP:journals/corr/abs-1711-02509}

\bibliography{ref}

\begin{thebibliography}{21}
\expandafter\ifx\csname natexlab\endcsname\relax\def\natexlab#1{#1}\fi

\bibitem[{Bunescu and Mooney(2005)}]{bunescu-mooney2005}
Razvan Bunescu and Raymond Mooney. 2005.
\newblock A shortest path dependency kernel for relation extraction.
\newblock In \emph{Proceedings of Human Language Technology Conference and
  Conference on Empirical Methods in Natural Language Processing}, pages
  724--731, Vancouver, British Columbia, Canada. Association for Computational
  Linguistics.

\bibitem[{Cai et~al.(2016)Cai, Zhang, and Wang}]{cai2016bidirectional}
Rui Cai, Xiaodong Zhang, and Houfeng Wang. 2016.
\newblock Bidirectional recurrent convolutional neural network for relation
  classification.
\newblock In \emph{ACL (1)}.

\bibitem[{Hendrickx et~al.(2010)Hendrickx, Kim, Kozareva, Nakov, S{\'e}aghdha,
  Pad\'{o}, Pennacchiotti, Romano, and Szpakowicz}]{Hendrickx2010}
Iris Hendrickx, Su~Nam Kim, Zornitsa Kozareva, Preslav Nakov, Diarmuid~\'{O}.
  S{\'e}aghdha, Sebastian Pad\'{o}, Marco Pennacchiotti, Lorenza Romano, and
  Stan Szpakowicz. 2010.
\newblock Semeval-2010 task 8: Multi-way classification of semantic relations
  between pairs of nominals.
\newblock In \emph{Proceedings of the 5th International Workshop on Semantic
  Evaluation}, SemEval '10, pages 33--38, Stroudsburg, PA, USA. Association for
  Computational Linguistics.

\bibitem[{Hochreiter and Schmidhuber(1997)}]{hochreiter1997long}
Sepp Hochreiter and J{\"u}rgen Schmidhuber. 1997.
\newblock Long short-term memory.
\newblock \emph{Neural computation}, 9(8):1735--1780.

\bibitem[{Liu et~al.(2017)Liu, Wang, Chang, and Sui}]{liu2017soft}
Tianyu Liu, Kexiang Wang, Baobao Chang, and Zhifang Sui. 2017.
\newblock A soft-label method for noise-tolerant distantly supervised relation
  extraction.
\newblock In \emph{Proceedings of the 2017 Conference on Empirical Methods in
  Natural Language Processing}, pages 1790--1795.

\bibitem[{Liu et~al.(2015)Liu, Wei, Li, Ji, Zhou, and Wang}]{liu2015dependency}
Yang Liu, Furu Wei, Sujian Li, Heng Ji, Ming Zhou, and Houfeng Wang. 2015.
\newblock A dependency-based neural network for relation classification.
\newblock In \emph{In Proceedings of the 53rd Annual Meeting of the Association
  for Computational Linguistics and the 7th International Joint Conference on
  Natural Language Processing (Volume 2: Short Papers}. Citeseer.

\bibitem[{Mikolov et~al.(2013)Mikolov, Sutskever, Chen, Corrado, and
  Dean}]{mikolov2013distributed}
Tomas Mikolov, Ilya Sutskever, Kai Chen, Greg~S Corrado, and Jeff Dean. 2013.
\newblock Distributed representations of words and phrases and their
  compositionality.
\newblock In \emph{Advances in neural information processing systems}, pages
  3111--3119.

\bibitem[{Santos et~al.()Santos, Xiang, and Zhou}]{santos2015classifying}
Cicero Nogueira~dos Santos, Bing Xiang, and Bowen Zhou.
\newblock Classifying relations by ranking with convolutional neural networks.

\bibitem[{Socher et~al.(2011)Socher, Pennington, Huang, Ng, and
  Manning}]{socher2011semi}
Richard Socher, Jeffrey Pennington, Eric~H Huang, Andrew~Y Ng, and
  Christopher~D Manning. 2011.
\newblock Semi-supervised recursive autoencoders for predicting sentiment
  distributions.
\newblock In \emph{Proceedings of the conference on empirical methods in
  natural language processing}, pages 151--161. Association for Computational
  Linguistics.

\bibitem[{Sun(2014{\natexlab{a}})}]{sun2014structure}
Xu~Sun. 2014{\natexlab{a}}.
\newblock Structure regularization for structured prediction.
\newblock In \emph{Advances in Neural Information Processing Systems}, pages
  2402--2410.

\bibitem[{Sun(2014{\natexlab{b}})}]{DBLP:journals/corr/Sun14c}
Xu~Sun. 2014{\natexlab{b}}.
\newblock Structure regularization for structured prediction: Theories and
  experiments.
\newblock \emph{CoRR}, abs/1411.6243.

\bibitem[{Sun et~al.(2017{\natexlab{a}})Sun, Ren, Ma, Wei, Li, and
  Wang}]{DBLP:journals/corr/abs-1711-06528}
Xu~Sun, Xuancheng Ren, Shuming Ma, Bingzhen Wei, Wei Li, and Houfeng Wang.
  2017{\natexlab{a}}.
\newblock Training simplification and model simplification for deep learning:
  {A} minimal effort back propagation method.
\newblock \emph{CoRR}, abs/1711.06528.

\bibitem[{Sun et~al.(2017{\natexlab{b}})Sun, Sun, Ma, Ren, Zhang, Li, and
  Wang}]{DBLP:journals/corr/abs-1711-10331}
Xu~Sun, Weiwei Sun, Shuming Ma, Xuancheng Ren, Yi~Zhang, Wenjie Li, and Houfeng
  Wang. 2017{\natexlab{b}}.
\newblock Complex structure leads to overfitting: {A} structure regularization
  decoding method for natural language processing.
\newblock \emph{CoRR}, abs/1711.10331.

\bibitem[{Wang et~al.(2016)Wang, Cao, de~Melo, and Liu}]{wang2016relation}
Linlin Wang, Zhu Cao, Gerard de~Melo, and Zhiyuan Liu. 2016.
\newblock Relation classification via multi-level attention cnns.
\newblock In \emph{ACL (1)}.

\bibitem[{Wang et~al.(2017)Wang, Li, Yang, Sun, and
  Wang}]{DBLP:conf/ijcnlp/WangLYSW17}
Yizhong Wang, Sujian Li, Jingfeng Yang, Xu~Sun, and Houfeng Wang. 2017.
\newblock Tag-enhanced tree-structured neural networks for implicit discourse
  relation classification.
\newblock In \emph{Proceedings of the Eighth International Joint Conference on
  Natural Language Processing, {IJCNLP} 2017, Taipei, Taiwan, November 27 -
  December 1, 2017 - Volume 1: Long Papers}, pages 496--505.

\bibitem[{Wen(2017)}]{DBLP:journals/corr/abs-1711-02509}
Ji~Wen. 2017.
\newblock Structure regularized bidirectional recurrent convolutional neural
  network for relation classification.
\newblock \emph{CoRR}, abs/1711.02509.

\bibitem[{Xu et~al.(2017)Xu, Wen, Sun, and
  Su}]{DBLP:journals/corr/abs-1711-07010}
Jingjing Xu, Ji~Wen, Xu~Sun, and Qi~Su. 2017.
\newblock A discourse-level named entity recognition and relation extraction
  dataset for chinese literature text.
\newblock \emph{CoRR}, abs/1711.07010.

\bibitem[{Xu et~al.(2015)Xu, Mou, Li, Chen, Peng, and Jin}]{xu2015classifying}
Yan Xu, Lili Mou, Ge~Li, Yunchuan Chen, Hao Peng, and Zhi Jin. 2015.
\newblock Classifying relations via long short term memory networks along
  shortest dependency paths.
\newblock In \emph{EMNLP}, pages 1785--1794.

\bibitem[{Zeiler()}]{zeiler2012adadelta}
Matthew~D Zeiler.
\newblock Adadelta: An adaptive learning rate method.

\bibitem[{Zeng et~al.(2014)Zeng, Liu, Lai, Zhou, Zhao
  et~al.}]{zeng2014relation}
Daojian Zeng, Kang Liu, Siwei Lai, Guangyou Zhou, Jun Zhao, et~al. 2014.
\newblock Relation classification via convolutional deep neural network.
\newblock In \emph{COLING}, pages 2335--2344.

\bibitem[{Zhang et~al.(2015)Zhang, Zheng, Hu, and
  Yang}]{zhang2015bidirectional}
Shu Zhang, Dequan Zheng, Xinchen Hu, and Ming Yang. 2015.
\newblock Bidirectional long short-term memory networks for relation
  classification.
\newblock In \emph{PACLIC}.

\end{thebibliography}
\bibliographystyle{acl_natbib}
\end{CJK*}
\end{document}